\documentclass[]{spie}  

\usepackage{amsmath,amsfonts,amssymb}
\usepackage{graphicx}
\usepackage[colorlinks=true, allcolors=blue]{hyperref}

\title{Memorizing SAM: 3D Medical Segment Anything Model with Memorizing Transformer}

\author[a]{Xinyuan Shao}
\author[a]{Yiqing Shen}
\author[a]{Mathias Unberath}
\affil[a]{Johns Hopkins University, Baltimore, MD, USA}

\begin{document} 
\maketitle

\begin{abstract}
Segment Anything Models (SAMs) have gained increasing attention in medical image analysis due to their zero-shot generalization capability in segmenting objects of unseen classes and domains when provided with appropriate user prompts. 
Addressing this performance gap is important to fully leverage the pre-trained weights of SAMs, particularly in the domain of volumetric medical image segmentation, where accuracy is important but well-annotated 3D medical data for fine-tuning is limited.
In this work, we investigate whether introducing the memory mechanism as a plug-in, specifically the ability to memorize and recall internal representations of past inputs, can improve the performance of SAM with limited computation cost.
To this end, we propose Memorizing SAM, a novel 3D SAM architecture incorporating a memory Transformer as a plug-in.
Unlike conventional memorizing Transformers that save the internal representation during training or inference, our Memorizing SAM utilizes existing highly accurate internal representation as the memory source to ensure the quality of memory. 
We evaluate the performance of Memorizing SAM in 33 categories from the TotalSegmentator dataset, which indicates that Memorizing SAM can outperform state-of-the-art 3D SAM variant \textit{i}.\textit{e}., FastSAM3D with an average Dice increase of 11.36\,\% at the cost of only 4.38 millisecond increase in inference time. 
%
%
The source code is publicly available at  \url{https://github.com/swedfr/memorizingSAM}.
\end{abstract}


\section{Introduction}
\label{sec:intro}  
Medical image segmentation is important in numerous diagnosis- and prognosis-related tasks, including lesion localization, tissue characterization, and volume estimation~\cite{dora2017state,mirikharaji2023survey,wu2022swin,jiang2022deep}. 
Although deep learning models such as U-Net and variants \cite{UNet,ke2023clusterseg,he2023transnuseg,MedNext} improved performance on specific tasks, they are confined to narrow application scopes and datasets.
Recently, the introduction of the Segment Anything Model (SAM) \cite{sam} has initiated a paradigm shift towards prompt-based interactive segmentation, leveraging the inherent generalizability of foundation models to provide generalization ability across various segmentation tasks.
SAM's architecture comprises a pre-trained Vision Transformer (ViT) encoder \cite{vit}, a prompt encoder, and a lightweight mask decoder that facilitates multi-mask prediction. 
Trained on a large-scale dataset of over 1 billion masks and 11 million images, SAM demonstrates the ability to adapt to diverse segmentation in a zero-shot manner. 
However, the direct application of SAM to medical image segmentation tasks reveals a performance gap compared to those fully supervised models due to the complexity of medical images. 
Therefore, approaches like MedSAM \cite{medsam} and SAM-Med2D \cite{sammed2d} have fine-tuned the SAM model on various 2D medical datasets.
When applied to 3D volumetric data using slice-wise processing \cite{mazurowski2023segment,bui2023sam3d}, these 2D SAMs approaches underperform 3D fully supervised DL models due to the loss of information between slices and suffer from long inference times.
To tackle these issues, SAM-Med3D \cite{sammed3d} and FastSAM3D \cite{shen2024fastsam3d} introduced 3D counterparts of SAM's components.

Despite these advancements, the performance of 3D medical SAMs, while achieving an overall Dice score of 0.5, still falls short of clinical requirements \cite{shen2024fastsam3d}. 
Moreover, the generalizability of these 3D medical SAMs to unseen datasets remains lower than expected. 
To address these challenges, we propose Memorizing SAM, which integrates the concept of Memorizing Transformers \cite{wu2022memorizing} with SAM.
Memorizing Transformers\cite{wu2022memorizing} were originally developed for natural language processing to enhance understanding of long text sequences.
Afterward, MoViT \cite{movit} extends the Memorizing Transformers\cite{wu2022memorizing} into the vision domain \textit{i}.\textit{e}., a ViT with memorizing block, which demonstrates high performance for image classification without extra GPU computation cost.
By incorporating this extra memory component, we aim to improve the performance of 3D medical SAM models on both seen and unseen data.
A key advantage of our proposed method is that the memory is prepared before inference. 
To be more specific, unlike traditional Memorizing Transformers\cite{wu2022memorizing}, our Memorizing SAM does not require any additional cache to store memory during run-time, which thus has no extra memory usage during inference.
Instead, all memory is saved within the file system and only loaded into the cache when needed, optimizing computational resources and potentially reducing inference time.

The major contributions are three-fold. Firstly, we proposed Memorizing SAM, which incorporates a memory Transformer as a plug-in module into the existing SAM framework. 
Secondly, Memorizing SAM load \texttt{key}-\texttt{value} pairs  into cache only when needed therefore minimize additional memory usage during inference.
Thirdly, our approach demonstrates performance improvements over state-of-the-art 3D SAM variants, particularly FastSAM3D.
\begin{figure}[t!]
    \centering
    \includegraphics[width=\linewidth]{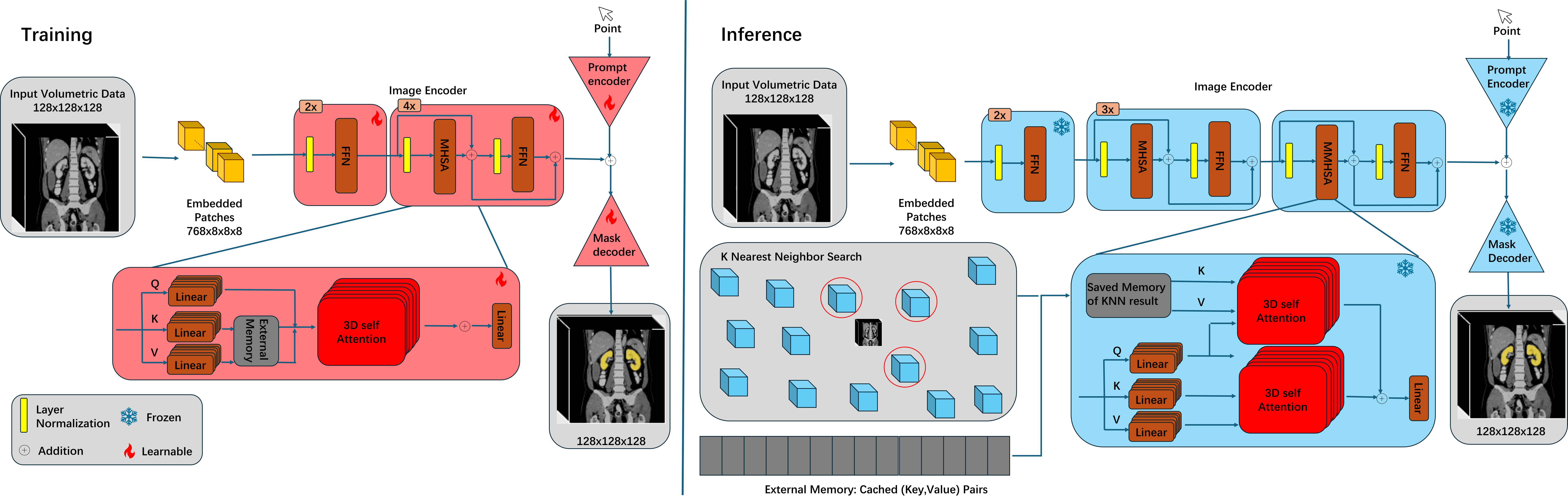}
    \caption{The overall architecture of Memorizing SAM, where we show the training and inference stages separately. 
    The training stage (left panel) is a general FastSAM3D training while saving the \texttt{key}, \texttt{value} pairs together with the current training image into external memory as a tuple. 
    The inference stage (right panel) incorporates \texttt{key}, \texttt{value} pairs from the external memory saved during the training stage via a kNN search.
    The memorizing block can be plugged into the Transformer block of any SAM or variants.
    }\label{fig:structure}
\end{figure}
\section{Methods}
\paragraph{Overview of Memorizing SAM}
We introduce Memorizing SAM, an enhanced 3D SAM model that incorporates an additional memory mechanism as a plug-in module as illustrated in Fig.~\ref{fig:structure}.
The training phase of Memorizing SAM begins with the selection of a high-quality dataset containing multi-class masks, whose quality will directly influence the efficacy of the generated memory. 
We then divide the selected dataset into $N$ one-class datasets, where $N$ represents the number of distinct object classes. 
Following this separation, we train $N$ separate Memorizing SAM models, each specialized for one of the $N$ object classes.
Using these trained class-specific SAM models, we process each one-class dataset to generate and save \texttt{key}-\texttt{value} pairs along with related model image inputs as tuples into the external memory for later use during the inference phase.
Unlike conventional Memorizing Transformers \cite{wu2022memorizing} that save internal representations during training or inference, our approach utilizes these pre-computed, highly accurate internal representations as the memory source, ensuring memory quality.
During the inference phase, Memorizing SAM leverages the previously saved memory to enhance its segmentation performance without incurring significant computational overhead.
Specifically, the memorizing block is integrated into the Transformer block of the SAM as a plug-in. 
This block incorporates \texttt{key}-\texttt{value} pairs from the external memory via a kNN search based on the image. 

\paragraph{Memorizing Transformer Block}
\label{Memorizing Transformer Block}
\label{block}
The memorizing block is designed as a plug-in module that can be integrated into any SAM variant, allowing for easy integration with different SAM variants. 
While preserving the standard dense self-attention and Feed-Forward Network (FFN) layers, our memorizing Transformer block introduces an innovative approximate kNN search mechanism that leverages a prepared high-quality memory.
Based on the dual use of queries for both local context and the prepared high-quality memory, the kNN lookup retrieves a set of memories, comprising the top-k (\texttt{key}, \texttt{value}) pairs that correspond to the current inference image query. 
The attention mechanism within the block operates as follows.
Initially, an attention matrix is constructed by computing the dot product between the query and the \texttt{key} from memory.
This matrix then undergoes softmax normalization.
The attention layer's output is derived by taking the dot product of this normalized attention matrix with the \texttt{value} from memory.
This process is iterated $k$ times, in contrast to the single iteration in standard dense attention. 
Simultaneously, attention to the local context is computed using conventional methods. 
To effectively combine the results of kNN-attention and local attention, we employ a ratio-based approach, where combination ratio is calculated by
$
\mathbf{R}_i = \frac{\mathbf{D}_i}{\sum_{i=0}^{k}{\frac{\mathbf{R}_L}{\mathbf{D}_i}}}$ ($i \in \{0,k\}
$), $\mathbf{A}_c = \mathbf{A}_L \cdot \mathbf{R}_L + \sum_{i=0}^k{\mathbf{R}_i \cdot \mathbf{A}_i}$.
Here $\mathbf{R}_L$ represents the preset ratio for local attention output $\mathbf{A}_L$, while $\mathbf{R}_i$ denotes the ratio for kNN-attention output $\mathbf{A}_i$, calculated using the kNN distance $\mathbf{D}_i$.
The parameter $k$ defines the scope of the kNN search, and $\mathbf{A}_c$ represents the final combined attention output.

\section{Experiments}
\begin{figure}[t!]
    \centering
    \includegraphics[width=\linewidth]{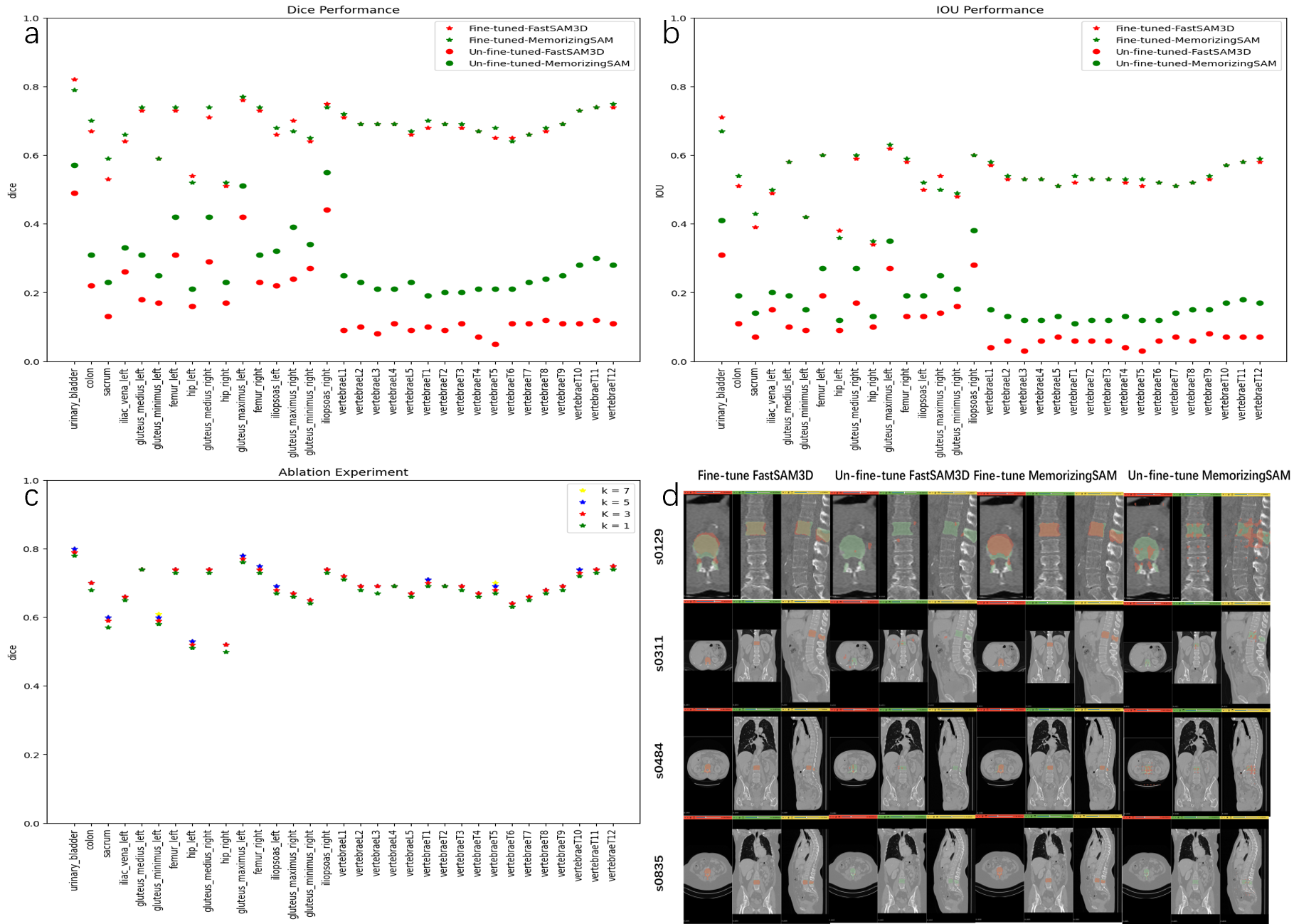}
    \caption{Performance comparison between the baseline 3D SAM model \textit{i}.\textit{e}., FastSAM3D and our proposed Memorizing SAM. 
    (a) Dice scores comparison across multiple anatomical classes, showing both fine-tuned and un-fine-tuned model versions of both models. 
    (b) Corresponding IoU (Intersection over Union) scores for the same experiments. 
    (c) Ablation study results demonstrating the impact of different number of $k$ in the Memorizing SAM. 
    (d) Visualization of segmentation results for representative samples, comparing fine-tuned and un-fine-tuned versions of both models across different anatomical structures. 
    These results highlight the improved performance of Memorizing SAM, particularly on challenging and underrepresented anatomical classes.
    }\label{fig:joint}
\end{figure}

\begin{table*}[!t]
\caption{
Computational Efficiency comparison between FastSAM3D and Memorizing SAM. 
The best performance is highlighted in \textbf{bold}.
Measurements are specific to the image encoder component in the SAM only.
}\label{table:compute}
\centering
\begin{tabular}{l|c|c|c|c}
\hline
\textbf{Model} & \textbf{Inference Time (ms)} & \textbf{FLOPs (G)} & \textbf{Memory (GB)} & \textbf{Parameters (M)} \\
\hline
\texttt{FastSAM3D} & \textbf{2.51} & \textbf{23.14} & \textbf{0.87} & \textbf{45.64} \\
\texttt{Memorizing SAM} & 6.89 & \textbf{23.14} & 0.88 & \textbf{45.64} \\
\hline
\end{tabular}
\end{table*}


\paragraph{Dataset}
We leverage TotalSegmentator dataset \cite{totalsegmentator} for performance evaluation. 
To better evaluate performance on a challenging and clinically relevant anatomical structure, we constructed a subset of TotalSegmentator that includes 33 classes from 500 samples for testing, 200 samples for SAM fine-tuning, and another 10 samples for tuning the Memorizing Transformer.
%
%

\paragraph{Results}
We set the hyper-parameter ratio $\mathbf{R}_L$ to 0.3 in the Memorizing SAM, to align with the previous work \cite{wu2022memorizing} which suggested that lower ratios for local memory tend to be more effective.
The kNN search parameter was set to $3$ to balance between leveraging historical information and maintaining computational efficiency.
Fig.~\ref{fig:joint} presents the performance comparison of FastSAM3D \cite{shen2024fastsam3d} and our proposed Memorizing SAM.
Our results demonstrate that Memorizing SAM consistently outperforms FastSAM3D across most anatomical classes, with particularly marginal improvements in un-fine-tuned scenarios. 
Specifically, Memorizing SAM achieved an average Dice score increase of 11.36\.\% on test data compared to FastSAM3D. 
While both models benefit from fine-tuning, the performance gap between Memorizing SAM and FastSAM3D narrows in fine-tuned scenarios. 
This suggests that the memory mechanism is particularly beneficial when applied to SAM models that have not undergone task-specific fine-tuning, potentially enhancing their ability to generalize to new segmentation tasks.
Visual examples of segmentation results are provided in Fig.~\ref{fig:joint} (d).

\paragraph{Comparison on Computational Efficiency}
Table \ref{table:compute} presents a detailed comparison of computational efficiency between FastSAM3D and our proposed Memorizing SAM. All experiments were conducted on an NVIDIA T4 GPU to ensure consistent evaluation conditions.
Memorizing SAM shows a modest increase in inference time, from $2.51$ ms for FastSAM3D to $6.89$ ms, representing an additional $4.38$ ms per image. 
This increase is attributable to the computations required by the memorizing Transformer block. However, this 174\% increase in inference time should be contextualized with the 11.36\% improvement in Dice scores.
Notably, the FLOPs (floating point operations) remain constant at 23.14G for both models, indicating that our memory augmentation does not significantly increase the computational complexity of the core operations. 
Memory usage shows only a marginal increase from 0.87GB to 0.88GB, demonstrating the efficiency of our memory management strategy. 
The number of parameters remains unchanged at 45.64M, highlighting that our approach enhances performance without expanding the model size.
By loading key-value pairs into the cache only when needed, we minimize additional memory overhead while still leveraging the benefits of memorization.

\paragraph{Ablation Study}
\label{ablation}
Fig.~\ref{fig:joint} shows the ablation study for $k$ in the kNN search.
Our experiments reveal that the effect of varying k on the model's performance is relatively modest. 
Specifically, we tested $k$ values of 1, 3, 5, and 7. 
The results show only slight variations in Dice scores across different anatomical structures for these k values, with $k=3$ generally providing the best balance between performance and computational efficiency.
This limited impact of higher k values suggests that the \texttt{key} and \texttt{value} pairs retrieved later in the kNN search results (\textit{i}.\textit{e}., those with lower similarity) contribute less significantly to the final segmentation output. 

\section{Conclusion}
\label{Conclusion}
We propose Memorizing SAM to integrate the Memorizing Transformers\cite{wu2022memorizing} with 3D medical SAM. 
Memorizing SAM incorporates a memory Transformer as a plug-in module, leveraging high-quality, pre-computed internal representations to enhance segmentation capabilities of SAM.
Our approach demonstrated significant performance improvements over state-of-the-art 3D SAM variant, coming with minimal computational overhead.
Future work can explore ways to combine the benefits of memorizing mechanism with traditional fine-tuning approaches.

\bibliography{report} 
\bibliographystyle{spiebib} 

\end{document}